\documentclass[lettersize,journal]{IEEEtran}
\usepackage{amsmath,amsfonts}
\usepackage{algorithmic}
\usepackage{algorithm}
\usepackage{array}
\usepackage[caption=false,font=normalsize,labelfont=sf,textfont=sf]{subfig}
\usepackage{textcomp}
\usepackage{stfloats}
\usepackage{url}
\usepackage{verbatim}
\usepackage{graphicx}
\usepackage{cite}
\usepackage{paralist}
\usepackage{booktabs}
\usepackage{multirow}
\usepackage{multicol}
\usepackage{xcolor}
\usepackage{soul}
\usepackage{colortbl}
\usepackage{bm}
\usepackage[citecolor=blue, colorlinks]{hyperref}
\begin{document}

\title{Enhancing Low-light Light Field Images with \\A Deep Compensation Unfolding Network}

\author{Xianqiang Lyu, and Junhui Hou,~\IEEEmembership{Senior Member,~IEEE}
\thanks{This work was supported in part by Hong Kong Research
Grants Council under Grant 11218121, and in part by Hong
Kong Innovation and Technology Fund under Grant MHP/117/21. }
\thanks{The authors are with the Department of Computer Science, City University of Hong Kong, Hong Kong SAR. Email: xianqialv2-c@my.cityu.edu.hk; jh.hou@cityu.edu.hk}}

\markboth{}%
{Shell \MakeLowercase{\textit{et al.}}: A Sample Article Using IEEEtran.cls for IEEE Journals}

\maketitle

\begin{abstract}
This paper presents a novel and interpretable end-to-end learning framework, called the deep compensation unfolding network (DCUNet), for restoring light field (LF) images captured under low-light conditions. DCUNet is designed with a multi-stage architecture that mimics the optimization process of solving an inverse imaging problem in a data-driven fashion. The framework uses the intermediate enhanced result to estimate the illumination map, which is then employed in the unfolding process to produce a new enhanced result. Additionally, DCUNet includes a content-associated deep compensation module at each optimization stage to suppress noise and illumination map estimation errors. To properly mine and leverage the unique characteristics of LF images, this paper proposes a pseudo-explicit feature interaction module that comprehensively exploits redundant information in LF images. The experimental results on both simulated and real datasets demonstrate the superiority of our DCUNet over state-of-the-art methods, both qualitatively and quantitatively. Moreover, DCUNet preserves the essential geometric structure of enhanced LF images much better. The code is publicly available at \url{https://github.com/lyuxianqiang/LFLL-DCU}.

\end{abstract}

\begin{IEEEkeywords}
Light fields, low-light enhancement, deep learning, algorithm unrolling.
\end{IEEEkeywords}

\section{Introduction}
\IEEEPARstart{L}{ight} field (LF) images capture the appearance and geometry of 3D environments, enabling various applications, including image post-refocusing \cite{ng2006digital,fiss2014refocusing}, accurate depth estimation \cite{wanner2013variational,chen2018accurate}, and 3D scene reconstruction \cite{kim2013scene,guo2020accurate}. But, these applications are vulnerable to LF images taken in less-than-ideal lighting settings with severe noise and color distortions. 
Although a considerable number of low-light enhancement methods for single images 
\cite{Ren2019,xu2020learning,li2021learning,jiang2021enlightengan},
simply applying these methods to process low-light LF images, i.e., separately enhancing the sub-aperture images (SAIs) of an LF image one by one, 
may produce unsatisfactory performance, e.g., the inconsistent colors and geometric structures across SAIs, due to the lack of modeling of the unique relationship among SAIs (i.e., the LF parallax structure).   
To simultaneously process SAIs, alternatively, one can reorganize the SAIs of an LF image as a video, which is then fed into a typical low-light video enhancement method \cite{chen2019seeing,wang2021seeing} 
However, the relationship among SAIs is naturally different from that among frames of a video (i.e., object and camera motion), thus resulting in limited performance. 
Therefore, it is highly desired to investigate low-light enhancement methods tailored to LF images. 

\begin{figure}
\begin{center}
\includegraphics[width=0.95\linewidth]{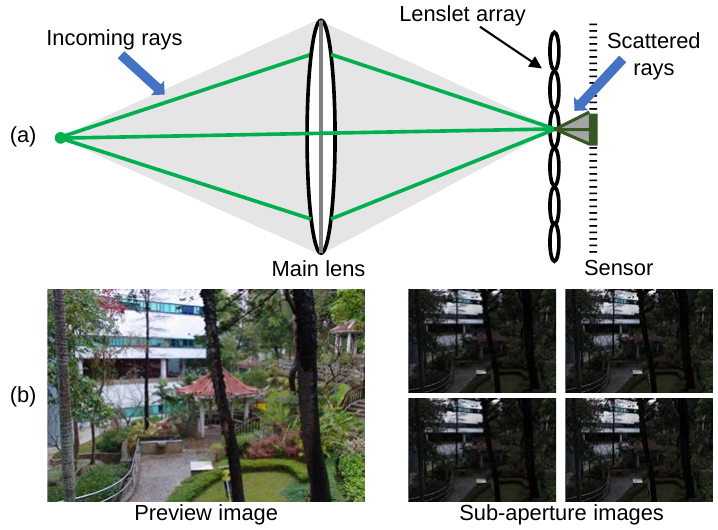} 
\vspace{-0.5cm}
\end{center}
   \caption{(a) Two key factors affecting brightness in the LF imaging process: incoming light intensity and ray scattering by the lenslet array. (b) An example illustrating the reduction in brightness in sub-aperture images as a consequence of scattered rays, as compared to the Lytro Illum camera preview.} 
\label{fig:rays}
\end{figure}

Low-light enhancement for LF images is still preliminary, and only a few methods have been proposed. The existing methods \cite{lamba2020harnessing,zhang2021effective,wang2023multi,wang2024pmsnet} are mainly focused on exploiting redundant information in LF images by improving the spatial-angular separable (SAS) convolution \cite{zhang2021effective} or utilizing predefined view stacks \cite{lamba2020harnessing,wang2023multi,wang2024pmsnet}. However, these methods cannot use the LF properties well, thus limiting their performance.
For instance, the neighboring SAI stacks used in \cite{lamba2020harnessing,wang2023multi} and selected SAIs in \cite{wang2024pmsnet} can only perceive the local information of the LF and inevitably suffer from the aliasing problem and artifacts when processing large disparity LF. Besides, they are purely data-driven, limiting their generalizability to some extent. 
Moreover, intense noise in low-light LF images will destroy the structural coherence of the LF and reduce the signal-to-noise ratio in extremely dark situations, which puts forward higher requirements for feature extraction.

In this paper, we tackle this challenging problem by proposing a deep compensation unfolding network (DCUNet), guided by the physical model of low-light LF imaging and the constrained optimization process of solving an inverse problem with the half quadratic splitting algorithm \cite{dong2018denoising,zhang2020deep}. Specifically, DCUNet is constructed with a multi-stage architecture that uses a deep optimization module to predict the enhancement results by leveraging the estimated illumination map and deep compensation for noise and illumination estimation error at each stage. To utilize the unique characteristic of LF images, i.e., the repeated observations (SAIs) of an identical scene, we propose a pseudo-explicit feature interaction module that emphasizes the traits of the epipolar plane images (EPIs) and converges rays from SAIs. Moreover, DCUNet emphasizes the necessity of integrating noise impact within the optimization process to enhance low-light LF with severe noise.

In summary, the main contributions of this paper are:
\begin{compactitem}
    \item a novel deep unfolding network for enhancing low-light LF images, featured with illumination map estimation, content-associated deep compensation, and deep optimization; 
    \item a purpose-engineered pseudo-explicit feature interaction module for comprehensively modeling information within and across views; and
    \item current state-of-the-art performance on both synthetic and real LF datasets with severe noise and extremely low light conditions.
\end{compactitem}

The remainder of this paper is organized as follows. In Section \ref{sec:RW}, we briefly review the related work. In Section \ref{sec:Problem}, we formulate the imaging process of low-light situations. In Section \ref{sec:Method}, we propose the architecture of our DCUNet in detail. Experimental results are presented in Section \ref{sec:Experiments}. Finally, we conclude this paper in Section \ref{sec:Conclusion}. 

\section{Related Work}
\label{sec:RW}

\subsection{Low-light Image/Video Enhancement}
As conventional low-light image enhancement methods, histogram equalization (HE)-based methods modify the histogram distribution of images to increase the dynamic range of an image at the global level \cite{histogram06} or local level \cite{Lee2013}. 
The Retinex theory \cite{Edwin1977}, which normally divides an image into reflectance and illumination, is also used in a number of other approaches \cite{Fu2016CVPR,guo2016lime,Structure2018}. Low-light image enhancement is treated as an illumination estimation problem since it is typically assumed that the reflectance component is constant regardless of the lighting.

With the emergence of deep learning, large amounts of data and deeper networks dramatically benefit low-light image enhancement tasks. 
Ren \textit{et al.} \cite{Ren2019} suggested a deep hybrid network for improving low-light images that simultaneously uses two streams to learn global information and salient structures. 
Xu \textit{et al.} \cite{xu2020learning} proposed a frequency-based decomposition and enhancement model for low-light image enhancement that recovered picture content at low frequency and details at high frequency.
EnlightenGAN \cite{jiang2021enlightengan}, an unsupervised GAN-based approach, learns to improve low-light images from unpaired low/normal light data.
Guo \textit{et al.} proposed a zero-reference curve estimation network (Zero-DCE) \cite{guo2020zero} and accelerated version (Zero-DCE++) \cite{li2021learning} that formulate the light enhancement as a task of image-specific curve estimating, which takes a low-light image as input and generates high-order curves. 
Ma \textit{et al.} \cite{ma2022toward} proposed a self-calibrated illumination learning framework for brightening images in real-world low-light scenarios. 
Zheng \textit{et al.} \cite{zheng2021adaptive} employed a deep unfolding total variation module to estimate the noise level map, which was then used to assist the subsequent image restoration module.
In contrast to existing techniques that rely on pixel-wise reconstruction losses and deterministic processes, Wang \textit{et al.} \cite{wang2022low} proposed a normalizing flow model trained with a negative log-likelihood loss, which takes low-light images or features as conditions. This approach naturally characterizes the structural context and measures the visual distance in the image manifold. 

Inspired by the Retinex model, Wei \textit{et al.} \cite{Chen2018Retinex} proposed the Retinex-Net deep network, which combines illumination mapping with image decomposition.
A network called KinD, which divides images into two parts, was developed by Zhang \textit{et al.} \cite{zhang2019kindling} to kindle the darkness of an image. The reflectance component removes deterioration, whereas the illumination component controls the light intensity.
Recently, various retinex-based deep unfolding frameworks for illumination estimation and image enhancement have been proposed \cite{liu2021retinex,wu2022uretinex}. Our method differs from them in three main aspects:  
$(\textbf{1})$ Instead of denoising as a separate module or preprocessing procedure, we take noise into account throughout the deep unfolding process;
$(\textbf{2})$ we alternately optimize the illuminance map and enhanced results to increase the information interaction at different stages; 
$(\textbf{3})$ we employ the pseudo-explicit feature interaction as the basic module of the network to adjust to the multi-view property of low-light LFs.

Low-light video enhancement methods \cite{lv2018mbllen,wang2019enhancing,chen2019seeing,triantafyllidou2020low,wang2021seeing,peng2022lve} have received increasing attention in recent years. Wang \textit{et al.} \cite{wang2019enhancing} explored the physical model using an LSTM-based network to enhance low-light videos.
Chen \textit{et al.} \cite{chen2019seeing} trained a dual-stream low-light video enhancement network on the released DRV dataset. 
Wang \textit{et al.} \cite{wang2021seeing} proposed an end-to-end method on the released SDSD dataset to achieve low-light enhancement and noise reduction simultaneously. 

\subsection{Low-light LF Image Enhancement}
Existing light-field low-light enhancement approaches tend to address data containing harsh noise in extremely dark environments. These approaches require enhancing brightness while maintaining view consistency across SAIs. Lamba \textit{et al.} \cite{lamba2020harnessing} developed a low-light L3F dataset and a two-stage L3Fnet architecture, which includes a global representation block to encode the LF geometry and a view reconstruction block to restore each view. To restore dark LFs efficiently, they propose a three-stage network \cite{lamba2022fast} that utilizes global, local, and view-specific information in low-light LFs and fuses them using an RNN-inspired feedforward network. 
Zhang \textit{et al.} \cite{zhang2021effective} extended the Retinex-Net to LF, developing a learning-based decomposition-enhancement method to decouple the complex task into several sub-tasks.
Lamba \textit{et al.} \cite{Lamba2022f} proposed a three-stage network that utilizes global, local, and view-specific information for fast and efficient low-light restoration in an RNN-inspired network. Zhang \textit{et al.} \cite{zhang2023lrt} utilized the Transformer for efficient low-light  LF image restoration.
Wang \textit{et al.} \cite{wang2023multi} proposed a multiple-stream progressive restoration network for low-light LF enhancement.
Wang \textit{et al.} \cite{wang2024pmsnet} proposed a parallel multi-scale network (PMSNet) that employs an Unet-like architecture to process light field features at different scales. The PMSNet network leverages both local and global information to achieve improved enhancement results, similar to the approach adopted by L3FNet \cite{lamba2020harnessing}.
These algorithms are mostly driven by data, and the network architecture is based on experience, which is not easily interpretable. Additionally, the LF's benefits for enhancing brightness are not fully utilized.

\begin{figure*}
\begin{center}
\includegraphics[width=0.9\linewidth]{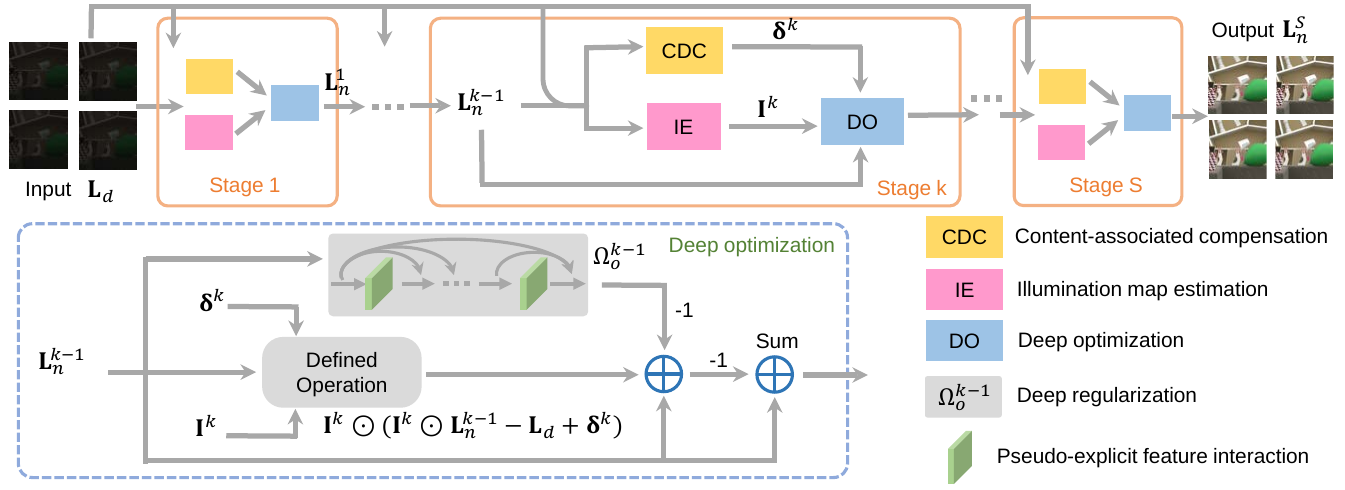}
\vspace{-0.4cm}
\end{center}
   \caption{Flowchart of the proposed end-to-end learning-based framework for low-light LF image enhancement, namely DCUNet. DCUNet consists of three main modules: Illumination Map Estimation (IE), Content-associated Deep Compensation (CDC), and Deep Optimization (DO). The CDC module compensates for the solution bias of the enhanced intermediate results $\textbf{L}_n^{k-1}$ and estimated illumination map $\textbf{I}^{k}$. The DO module is designed to mimic the optimization process and iteratively refine the enhanced result.
   }
\label{fig:pipeline}
\end{figure*}

\section{Problem Formulation}
\label{sec:Problem}

In low-light situations, the imaging process produces a variety of loud noises, including shot noise, dark current, and readout noise. The readout noise follows a Gaussian distribution, while the shot noise and dark current both follow Poisson distributions \cite{wei2020physics,remez2018class}. As Poissonian-Gaussian noise can be formulated as an additive \textit{signal-dependent} noise through heteroskedastic normal approximation \cite{foi2008practical}, we thus formulate the LF imaging process under low-light conditions as 
\begin{equation}
    \textbf{L}_{d} =  \alpha \textbf{K} \odot \textbf{L}_{n} + \textbf{N} ,
    \label{eqution1}
\end{equation}
where $\textbf{L}_{n}\in \mathbb{R}^{u \times v \times c \times h \times w}$ (with angular resolution $u \times v$, channel number $c$, and spatial resolution $h \times w$) is the LF image with normal light, $\alpha$ is the photoelectron amplification factor proportional to the scene illumination, $\textbf{K} \in \mathbb{R}^{u \times v \times c \times h \times w}$ is the system gain composed of analog and digital gain, $\textbf{N} \in \mathbb{R}^{u \times v \times c \times h \times w}$ is the mixture noise composed of \textit{signal-independent} Gaussian noise and \textit{signal-dependent} Poisson noise, $\textbf{L}_{d}\in \mathbb{R}^{u \times v \times c \times h \times w}$ is the observed low-light LF image, and the operator $\odot$ means element-wise multiplication. Based on the Retinex theory \cite{Edwin1977}, we can use the illumination map $\textbf{I} \in \mathbb{R}^{u \times v \times c \times h \times w}$ to represent the $\alpha \textbf{K}$, and thus, rewrite Eq. \eqref{eqution1} as 
\begin{equation}
    \textbf{L}_{d} = \textbf{I} \odot \textbf{L}_{n} + \textbf{N}.
    \label{eqution2}
\end{equation}

To address the ill-posed problem of recovering $\textbf{L}_{n}$ from $\textbf{L}_{d}$ in Eq. \eqref{eqution2}, we formulate a constrained optimization problem by introducing a regularization term $ \mathcal{R}(\cdot)$ to constrain the solution space of $\textbf{L}_{n}$: 
\begin{equation}
    \begin{split}
        \min_{\textbf{L}_{n},\bm{\nu},\textbf{I},\textbf{N}} & \left \| \textbf{L}_{d} - \textbf{I} \odot \textbf{L}_{n} - \textbf{N} \right \|_{2}^{2} +\lambda \mathcal{R} (\bm{\nu} ) \\
        & s.t. \quad \textbf{L}_{n}=\bm{\nu} .
    \end{split}
    \label{eqution3}
\end{equation}
where $\lambda$ is a positive penalty parameter, and the auxiliary variable $\bm{\nu}$ is used to facilitate optimization \cite{venkatakrishnan2013plug}. It is worth noting that there are two key differences between this optimization problem and the general optimization process. First, Eq. \eqref{eqution3} takes into account the impact of strong noise on the optimization process by including the noise term $\textbf{N}$ in the objective function. Second, the illumination map $\textbf{I}$ and noise term $\textbf{N}$ are considered as \textit{signal-dependent} variables that are related to the optimization result $\textbf{L}_{n}$ rather than free variables. 

Consequently, we can leverage the $\textbf{L}_{n}^{k-1}$ to calculate the $\textbf{I}^{k}$ and $\textbf{N}^{k}$, and employ the half-quadratic splitting method \cite{dong2018denoising,zhang2020deep} to solve the problem in Eq. \eqref{eqution3} by alternatively and iteratively tackling the following two sub-problems:
\begin{equation}
    \small
    \begin{cases}
    \textbf{L}_{n}^{k} = \underset{\textbf{L}_{n}}{\texttt{argmin}} \left \| \textbf{L}_{d} - \textbf{I}^{k} \odot \textbf{L}_{n} - \textbf{N}^{k} \right \|_{2}^{2} + \mu  \left \| \textbf{L}_{n} - \bm{\nu} ^{k-1}  \right \|_{2}^{2} \\
    \bm{\nu} ^{k} = \underset{\bm{\nu} }{\texttt{argmin}} \; \mu \left \| \textbf{L}_{n}^{k} - \bm{\nu}  \right \|_{2}^{2} + \lambda \mathcal{R} (\bm{\nu} ) ,
    \end{cases}
    \label{eqution4}
\end{equation}
where $k$ is the iteration index, $\mu>0$ is the penalty parameter. With the estimated $\textbf{I}^{k}$, $\textbf{N}^{k}$, and a deep regularization term, we can obtain the enhancement result by solving Eq. \eqref{eqution4}.

\section{Proposed Method}
\label{sec:Method}

\noindent\textbf{Overview}.  As shown in Fig. \ref{fig:pipeline}, we propose a deep unfolding network to achieve low-light LF image enhancement end-to-end, namely DCUNet, which mimics the optimization process in Eq. \eqref{eqution4}. 
DCUNet, constructed with a multi-stage architecture, is mainly composed of three modules, i.e., illumination map estimation, content-associated deep compensation, and a deep optimization module. Specifically, we first use the result of the preceding iteration to estimate the illumination map. However, the estimation of the illumination map may contain errors, and the Poisson and Gaussian noise cannot be ignored, which can negatively affect the optimization process. Thus, we suppress the noise and illumination map estimation error by content-associated deep compensation in each optimization stage. Then, we adopt a deep optimization module to map the update strategy using a deep regularization term and a specified calculation. In what follows, we will detail each module.

\begin{figure}
\begin{center}
\includegraphics[width=0.95\linewidth]{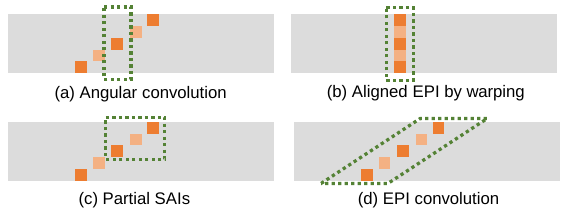} 
\vspace{-0.3cm}
\end{center}
   \caption{Demonstration of the utilization of redundant information by different strategies in an epipolar plane image (EPI) perspective. The orange blocks with different saturations represent corresponding pixels affected by severe noise, and the green bounding boxes show the scope of different operations.
   }
\label{fig:epi}
\end{figure}

\subsection{Pseudo-explicit Feature Interaction}
Low-light conditions and microlens diffraction in LF imaging reduce ray intensity, as seen in Fig. \ref{fig:rays}(a), where scattered rays are dimmer than focused rays. Fig. \ref{fig:rays}(b) further shows that Lytro Illum's decoded sub-aperture images are less bright than the camera's preview.
Therefore, converging rays can brighten the LF to some extent. Besides, repeated observations from different viewpoints of the scene are helpful for denoising and brightness enhancement. The depth-guided warping is a direct and explicit way to achieve information interaction between SAIs. However, it is challenging to obtain accurate depth information for low-light LF images with severe noise. Therefore, we propose this module to achieve pseudo-explicit feature interaction without depth information by converging rays and emphasizing the importance of EPI convolution.\\

\begin{figure}
\begin{center}
\includegraphics[width=0.9\linewidth]{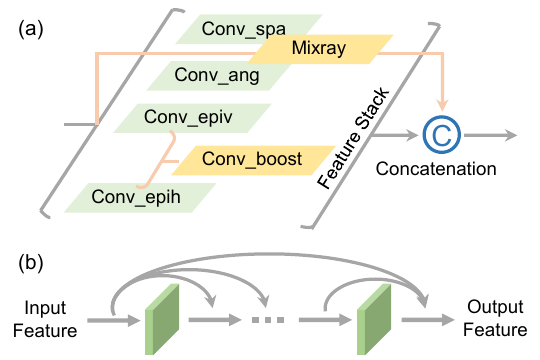}
\vspace{-0.3cm}
\end{center}
   \caption{Illustration of (\textbf{a}) the pseudo-explicit feature interaction module and (\textbf{b}) the corresponding dense-skip connection.}
\label{fig:feature}
\end{figure}

\noindent{\textbf{Analysis of different feature extraction strategies.}} 
Yeung \textit{et al.} \cite{yeung2018light} proposed the spatial-angular separable (SAS) convolution, which builds upon the general spatial convolution by using spatial convolution to extract SAI's features and angular convolution (as shown in Fig. \ref{fig:epi}(a)) to extract angular features. The angular convolution associates the information of different SAIs through multiple convolution operations in an indirect way, which may not be effective in handling LF images with large disparity and severe noise.
Fig. \ref{fig:epi}(b) shows the aligned EPI line via the depth-based warping operation, which can be directly fused/averaged to improve brightness and suppress noise. 
Fig. \ref{fig:epi}(c) uses the neighboring SAIs of the target SAI to assist feature extraction \cite{lamba2020harnessing,wang2023multi}. However, only local redundant information can be exploited. 
As depicted in Fig. \ref{fig:epi}(d), the direct processing of EPI information \cite{wu2017light,wang2022disentangling,cheng2022spatial} represents a highly effective approach for addressing large disparity low-light LFs. 
Therefore, in addition to using EPI convolution, we need to further emphasize the crucial significance of EPI information.\\

\begin{figure*}
\begin{center}
\includegraphics[width=0.95\linewidth]{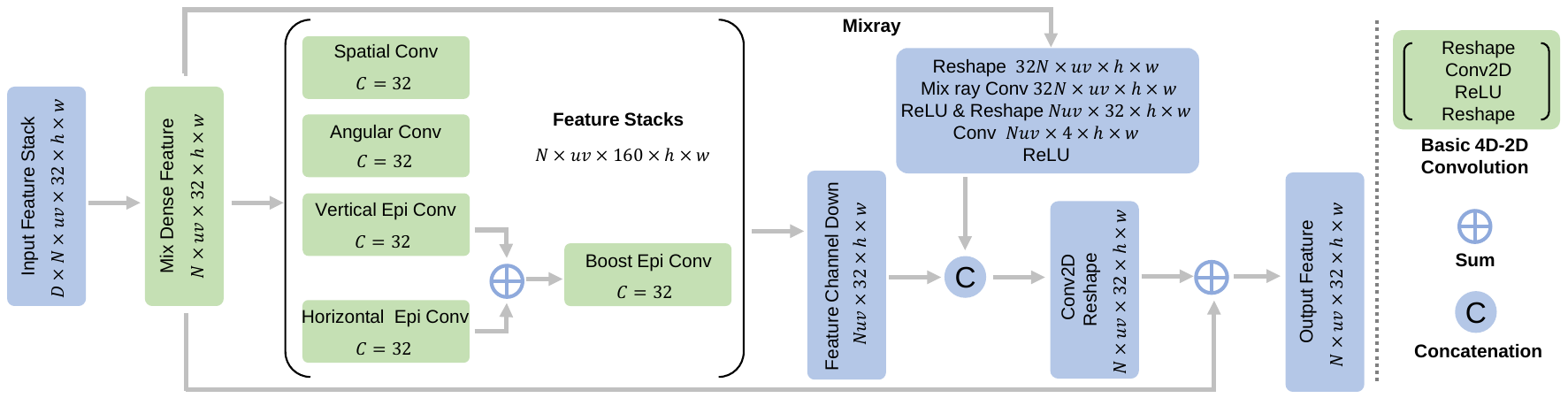}
\vspace{-0.3cm}
\end{center}
   \caption{Detail architecture of the proposed pseudo-explicit feature (DPEF) interaction module. The input of the DPEF encompasses dense feature stack (spatial, angular, and epipolar plane image) information, with stack parameter $D\in \left \{ 1,..., L \right \}$, where $L$ represents the layer number. The kernel size is 3 and the stride is 1 for all the basic 4D-2D convolutions. To balance the quality of the low-light LF enhancement and computational expenses, we ultimately settled on 6 layers, and 32 channel numbers in our framework.}
\label{fig:details-dpef}
\end{figure*}

\noindent \textbf{Module details.} As shown in Fig. \ref{fig:feature}, the proposed pseudo-explicit feature interaction module includes feature boost convolution and ray fusion, in addition to the four kinds of 2D convolution applied to different dimensions of an LF, i.e., spatial, angular, horizontal EPI, and vertical EPI. Specifically, we fuse horizontal $f_{epih}$ and  vertical $f_{epiv}$ EPI features using a convolution layer $\texttt{Conv}(\cdot)$ to boost the EPI information, i.e.,
\begin{equation}
    f_{boost}=\texttt{Conv}(f_{epiv} + f_{epih}).
    \label{eqution5}
\end{equation}
The ray fusion component mixes the features of all $N$ SAIs in the input features $f_{in}$ to simulate the process of ray convergence in the microlens, i.e.,
\begin{equation}
    f_{mix}=\texttt{Conv}(\sum_{i=1}^{N} \beta_{i} f_{in}^{i}),
    \label{eqution6}
\end{equation}
where $f_{mix}$ denotes the aggregated ray feature, and $\beta_{i}$ is a learnable fusion weight, which is achieved using the mix ray convolution layer in Fig. \ref{fig:details-dpef}.  After reducing the feature channel, the aggregated ray feature is concatenated with mixed results of five partially convolution features. The detailed architecture of the pseudo-explicit feature interaction module is shown in Fig. \ref{fig:details-dpef}.
Besides, we also use a dense-skip connection between pseudo-explicit feature interaction modules as Fig. \ref{fig:feature}(b) when performing feature extraction.

\subsection{Illumination Map Estimation}
The illumination map plays a crucial role \cite{Chen2018Retinex,wu2022uretinex,zhang2023lrt} for Retinex-based low-light image enhancement. In the deep unfolding process of \cite{wu2022uretinex}, the illumination map and the enhancement result are optimized simultaneously as independent variables. However, our method takes a different approach by estimating the illumination map using the enhancement results.
Specifically, at the $k$-th stage, based on the enhanced result of the preceding stage $\textbf{L}_{n}^{k-1}$ and the input low-light LF image $\textbf{L}_{d}$, we compute a coarse illumination map $\tilde{\textbf{I}}^{k}$, i.e., 
\begin{equation}
    \tilde{\textbf{I}}^{k} =\texttt{max}_{c \in {\{R,G,B \}}} (\textbf{L}_{n}^{k-1}-\gamma (\textbf{L}_{n}^{k-1}-\textbf{L}_{d}))_c,
    \label{eqution8}
\end{equation}
where $\gamma$ is a learnable parameter that adaptively suppresses the influence of the brightness mutation region. The $ \texttt{max}_{c \in {\{R,G,B \}}}$ operation \cite{land1977retinex,guo2016lime} can obtain global brightness while adeptly mitigating the effects of non-uniform illuminations, under the assumption that the illumination at any given spatial coordinate should not be inferior to the maximal intensity value among the \{R,G,B\} channels.

We further solve the following minimization problem to obtain the expected illumination map $\textbf{I}^{k}$:
\begin{equation}
    \min_{\textbf{I}^{k}} \frac{1}{2} \left \| \textbf{I}^{k}-\tilde{\textbf{I}}^{k} \right \| _{2}^{2}+ \mathcal{R}(\textbf{I}^{k}),
    \label{eqution9}
\end{equation}
where the $\mathcal{R}(\cdot)$ represents a regularization term of $\textbf{I}^{k}$. The problem in Eq. \eqref{eqution9} has a closed-form solution: $\textbf{I}^{k}=\tilde{\textbf{I}}^{k}-\partial \mathcal{R}(\textbf{I}^{k})$. In our implementation, we employ a deep gradient descent module $\Psi_{i}^{k}(\cdot)$ to approximate the gradient of the regularization term $\partial \mathcal{R}(\textbf{I}^{k})$. 

To achieve more accurate gradient estimation, it is essential to have precise initial optimization points as inputs to the module. In Fig. \ref{fig:stages}, we demonstrate the optimization results of each stage, and it is evident that the enhancement results change significantly in each stage. Therefore, in comparison to the illumination map of the preceding stage $\textbf{I}^{k-1}$, $\tilde{\textbf{I}}^{k}$ is closer to the optimization goal and is a more suitable input for the deep gradient descent module $\Psi_{i}^{k}(\cdot)$.  Fig. \ref{fig:detail-framework} depicts the detailed architecture of this module.

\begin{figure}
\begin{center}
\includegraphics[width=1\linewidth]{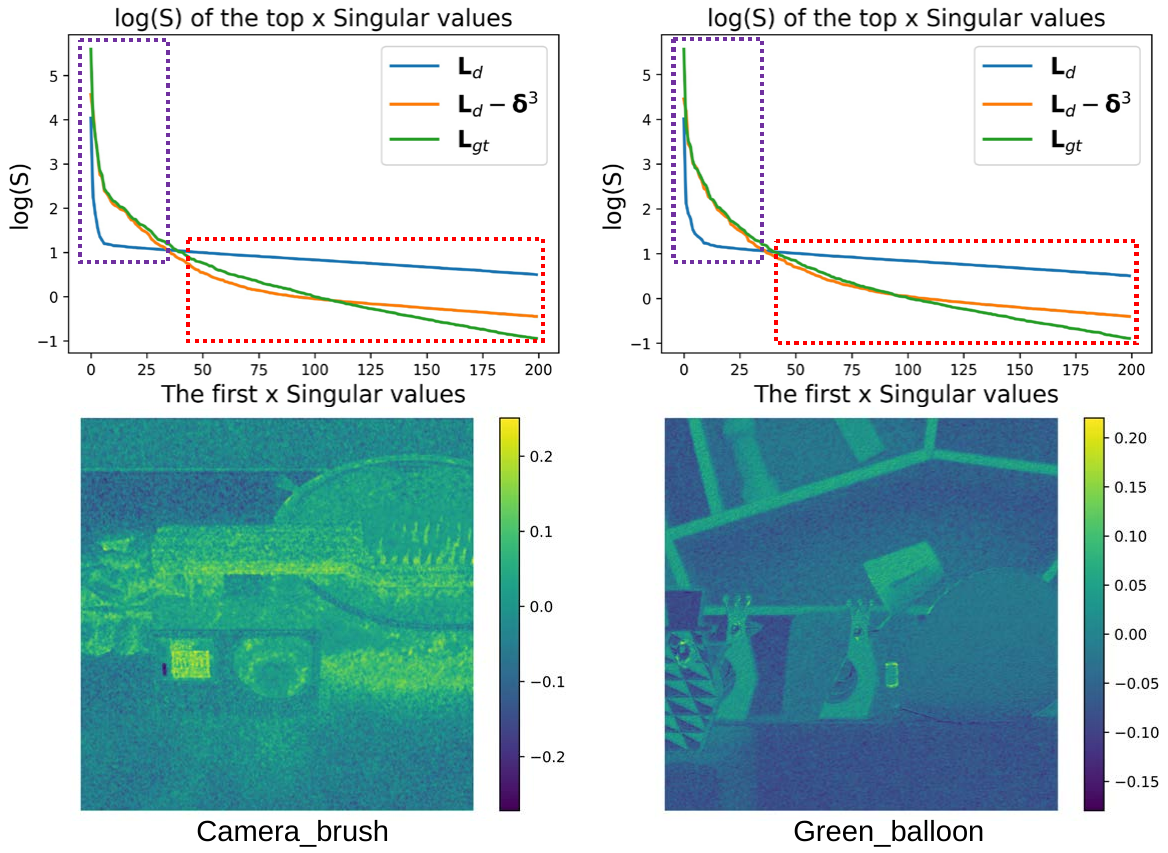}
\vspace{-0.7cm}
\end{center}
   \caption{
   Feature maps of content-associated deep compensation and singular values analysis for input low-light $\textbf{L}_d$, third stage compensated result $\textbf{L}_d - \bm{\delta}^3$, and ground truth $\textbf{L}_{gt}$. To make the comparison more obvious, the log results of the singular values are shown here. The singular values analysis results can be divided into purple and red regions, where the purple region is related to brightness compensation, and the red region is related to noise compensation.
   }
\label{fig:svd}
\end{figure}

\subsection{Content-associated Deep Compensation}
Low-light light field images with low signal-to-noise ratios pose significant challenges to enhancement algorithms \cite{guo2020zero,zheng2021adaptive}. In the context of the deep unfolding process, disregarding the noise term $\textbf{N}$ in Eq. \eqref{eqution3} introduces a non-negligible bias in the solution spaces of the intermediate results $L_n^{k-1}$ and the illumination map $I^k$ estimated from $L_n^{k-1}$. However, unsupervised removal of severe noise $\textbf{N}$ in low-light light field images through a single denoising process is a challenging task. Therefore, we propose a Content-associated Deep Compensation (CDC) module to mitigate the negative effects of noise and illumination map errors on the optimization process. By leveraging the CDC module's output $\bm{\delta}^k$, we approximate and replace the noise term $\textbf{N}^k$ in Eq. \eqref{eqution4}, thereby improving the robustness and accuracy of the optimization process.

\begin{figure*}
\begin{center}
\includegraphics[width=0.85\linewidth]{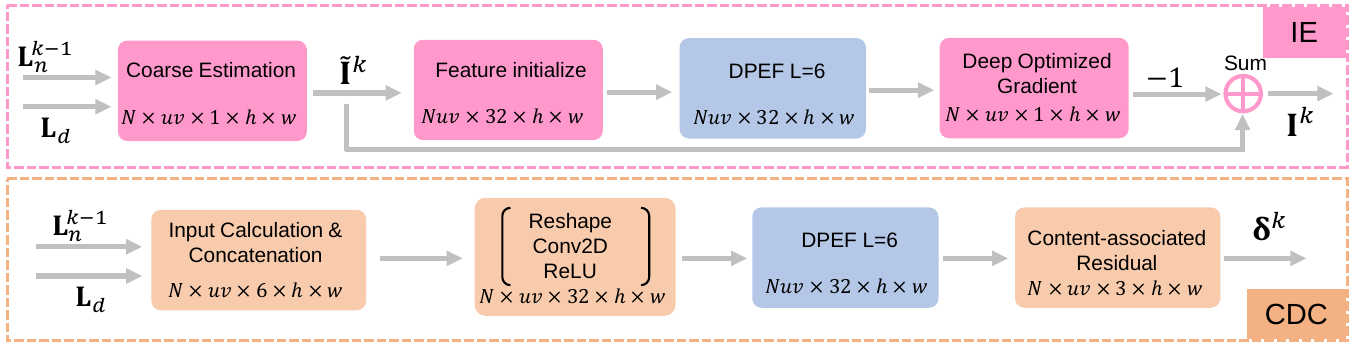}
\vspace{-0.4cm}
\end{center}
   \caption{Detailed architectures of the proposed illumination map estimation module (IE) and content-associated deep compensation (CDC).}
\label{fig:detail-framework}
\end{figure*}

Specifically, at the $k$-th stage, we feed the concatenation of  the enhanced result of the preceding stage $\textbf{L}_{n}^{k-1}$ and the input low-light LF image $\textbf{L}_{d}$ into a sub-network $\Phi_{c}^{k}(\cdot)$ for content-associated deep compensation, i.e.,  
\begin{equation}
    \bm{\delta}^{k}=\Phi_{c}^{k} \left(\texttt{Cat}\left(\textbf{L}_{d},\frac{\texttt{mean}(\textbf{L}_{d})}{\texttt{mean}(\textbf{L}_{n}^{k-1})}\textbf{L}_{n}^{k-1} \right)\right),
    \label{eqution7}
\end{equation}
where $\Phi_{c}^{k}(\cdot)$ is composed of pseudo-explicit feature interaction modules in stage $k$, and $\texttt{mean}(\cdot)$ returns the average intensity of its input. 
The detailed architecture of this module is shown in Fig. \ref{fig:detail-framework}.

To intuitively demonstrate the efficacy of content-associated deep compensation, we illustrate the feature maps of $\bm{\delta}^{k}$ and conduct a singular values analysis, as depicted in Fig. \ref{fig:svd}. The logarithmic results of the top 200 singular values for $\textbf{L}_d$, $\textbf{L}_d - \bm{\delta}^3$, and the ground truth LF image $\textbf{L}_{gt}$ demonstrate the module's capability in mitigating noise and compensating for errors in illumination map estimation.
As discussed in \cite{guo2015efficient}, the presence of small singular values can be attributed to noise. 
As shown in Fig. \ref{fig:svd}, the $\textbf{L}_d$ exhibits numerous singular values associated with noise in the red region. Conversely, the singular value of $\textbf{L}_d - \bm{\delta}^3$ in the red region is more similar to $\textbf{L}_{gt}$, thus highlighting the effectiveness of noise compensation.
Moreover, the singular values of $\textbf{L}_d - \bm{\delta}^3$ in the purple region lie between those of the low-light and ground truth images, demonstrating the ability of brightness compensation.

\subsection{Deep Optimization}
With the estimated illumination map $\textbf{I}^{k}$ and content-associated deep compensation $\bm{\delta}^{k}$ in the $k$-th iteration, we replace the $\textbf{N}^k$ in Eq. \eqref{eqution4} by $\bm{\delta}^{k}$, and address the optimization problem through a deep optimization module that comprises a defined computation and a deep regularization term. Specifically, we construct the deep implicit regularization term by utilizing pseudo-explicit feature interaction modules with a dense-skip connection. The solution can be expressed as
\begin{align}
    \begin{cases}
    \textbf{L}_{n}^{k} = \textbf{L}_{n}^{k-1} - \eta \left[\textbf{I}^{k} \odot \left(\textbf{I}^{k} \odot \textbf{L}_{n}^{k-1} - \textbf{L}_{d} + \bm{\delta}^{k}\right) \right. & \\
    \left. \qquad \qquad \qquad + \mu \left(\textbf{L}_{n}^{k-1} - \bm{\nu}^{k-1} \right) \right], & \\
    \bm{\nu}^{k} = \mathcal{D}(\textbf{L}_{n}^{k}), &
    \end{cases}
\end{align}
where the $\textbf{L}_{n}^{k}$ a single step of gradient descent for an inexact solution, $\eta$ is the parameter controlling the step size, $\mathcal{D}(\cdot)$ represents the proximal operator \cite{zhang2018ista} to approximate the solution of $\bm{\nu}^{k}$.
To efficiently convert the proposed solution into a CNN architecture, we define an essential computation operation $\textbf{I}^{k} \odot (\textbf{I}^{k} \odot \textbf{L}_{n}^{k-1} - \textbf{L}_{d} + \bm{\delta}^{k})$ and replace the proximal operator $\mathcal{D}(\textbf{L}_{n}^{k})$ 
with a deep regularization module $\Omega_{o}^{k}(\textbf{L}_{n}^{k})$. 
Note that the modules $\Psi_{i}^{k}(\cdot),~\Phi_{c}^{k}(\cdot),~\Omega_{o}^{k}(\cdot)$ do not share parameters across stages to ensure the learning and representation capability of the framework.

\subsection{Loss Function}
We train our method by minimizing the following loss function:
    \begin{align}
        \ell = & \lambda_{1} \left \| \textbf{L}_{n} - \textbf{L}_{gt} \right \| _{1} + \lambda_{2}(1-\texttt{SSIM}(\textbf{L}_{n},\textbf{L}_{gt}))  \nonumber \\
        & +\lambda_{3}\texttt{Per}(\textbf{L}_{n},\textbf{L}_{gt}),
    \end{align}
where $LF_{gt}$ is the ground truth LF image, $\texttt{SSIM}(\cdot)$ represents the structural similarity loss, $\texttt{Per}(\cdot)$ denotes the perceptual loss \cite{johnson2016perceptual} calculated from high-level features of the pre-trained VGG network on ImageNet, and the three hyperparameters $\lambda_{1}$, $\lambda_{2}$ and $\lambda_{3}$ are empirically set to 1, 1, and 0.1, respectively.

\begin{figure*}
\begin{center}
\includegraphics[width=0.95\linewidth]{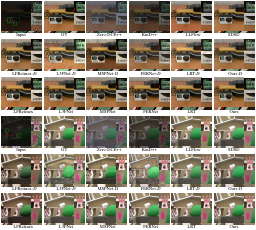}
\vspace{-0.3cm}
\end{center}
   \caption{Visual comparisons of enhanced center SAI by different methods on the synthetic LF dataset. The EPI and zoomed selected regions of the input image are the original images $\times 4$ to show the severe noise. It is recommended to view this figure by zooming in.}
\label{fig:synresults}
\end{figure*}

\begin{table}
\begin{center}
\caption{Quantitative comparisons (PSNR/SSIM/LPIPS) of different methods over the synthetic LF dataset. \textsc{Syn-F} and \textsc{Syn-D} represent the synthetic LF datasets of fixed and dynamic illumination environments, respectively. The best and second best results are highlighted in \textbf{bold} and \underline{underline}, respectively.} 
\vspace{-0.3cm}
\label{tab:my-table1}
\resizebox{\columnwidth}{!}{%
\begin{tabular}{c|c|ccc}
\toprule[1.2pt]
Dataset              & Method    & PSNR $\uparrow$ & SSIM $\uparrow$ & LPIPS $\downarrow$ \\ \hline\hline
\multirow{12}{*}{\textsc{Syn-F}} 
& \textbf{image \& video method} & & & \\ 
& ZeroDCE++ \cite{li2021learning}    & 11.31 & 0.132  & 1.123   \\
& KinD++ \cite{zhang2019kindling}    & 12.59 & 0.187  & 0.820    \\
& LLFlow \cite{wang2022low}    & 26.83 & 0.776  & 0.341    \\ 
& SDSD \cite{wang2021seeing}         & 19.07 & 0.463  & 0.686   \\ \cline{2-5}
& \textbf{LF image method} & & & \\ 
& LFRetinex \cite{zhang2021effective} & 25.72 & 0.762  & 0.225   \\
& L3Fnet \cite{lamba2020harnessing}  & 30.64 & 0.803  & 0.143 \\
& MSPnet \cite{wang2023multi}  & \underline{31.93} & \underline{0.860}  & \underline{0.131} \\
& FERNet \cite{Lamba2022f}  & 31.30 & 0.835  & 0.148 \\
& LRT \cite{zhang2023lrt}   & 28.09 & 0.768  & 0.259 \\
& Ours     & \textbf{33.96} & \textbf{0.886}  & \textbf{0.103}   \\ \hline
\multirow{7}{*}{\textsc{Syn-D}}
& \textbf{LF image method} & & & \\ 
& LFRetinex-\textit{D} & \underline{24.01} & 0.732  & 0.244   \\
& L3Fnet-\textit{D}    & 19.26 & 0.771  & 0.206   \\
& MSPnet-\textit{D}   & 21.98 & \underline{0.783}  & \underline{0.159}   \\
& FERNet-\textit{D}   & 21.85 & 0.716  & 0.221 \\
& LRT-\textit{D}      & 23.30 & 0.669  & 0.381 \\
& Ours-\textit{D}      & \textbf{26.90} & \textbf{0.806}  & \textbf{0.150}   \\ 
 \bottomrule[1.2pt]
\end{tabular}
}
\end{center}
\end{table}

\begin{table*}
\begin{center}
\caption{Quantitative comparisons (PSNR/SSIM/LPIPS ($\pm$ STD)) of different methods over the L3F dataset \cite{lamba2020harnessing}. The best and second best results are highlighted in \textbf{bold} and \underline{underline}, respectively. Note that the results of MSPnet* \cite{wang2023multi}, which demonstrate comparable performance to those reported in the original paper,  
were obtained by selecting the best model through testing multiple times across training iterations.  The results of MSPnet \cite{wang2023multi} were adopted for comparison in the following experiments to ensure a fair and unbiased evaluation.
}
\vspace{-0.4cm}
\label{tab:my-table2}
\resizebox{\textwidth}{!}{%
\begin{tabular}{c|ccc|ccc|ccc}
\toprule[1.2pt]
\multirow{2}{*}{method} & \multicolumn{3}{c|}{L3F-20} & \multicolumn{3}{c|}{L3F-50} & \multicolumn{3}{c}{L3F-100} \\ \cline{2-10}
 & PSNR $\uparrow$  & SSIM $\uparrow$   & LPIPS $\downarrow$   & PSNR $\uparrow$    & SSIM $\uparrow$   & LPIPS $\downarrow$   & PSNR $\uparrow$  & SSIM $\uparrow$   & LPIPS $\downarrow$   \\ \hline \hline
\textbf{image \& video method} & & & & & & & & & \\ 
ZeroDCE++ \cite{li2021learning} & 17.69 $\pm 1.74$  & 0.602 $\pm 0.13$ & 0.479 $\pm 0.09$  & 14.98 $\pm 2.20$  & 0.406 $\pm 0.11$ &  0.729 $\pm 0.10$  & 13.43 $\pm 1.73$  & 0.250 $\pm 0.10$  & 0.879 $\pm 0.10$  \\
KinD++ \cite{zhang2019kindling}   &  18.29 $\pm 2.88$ & 0.668 $\pm 0.12$ & 0.201 $\pm 0.08$  & 17.92 $\pm 2.50$  & 0.609 $\pm 0.09$ &  0.268 $\pm 0.08$ & 17.37 $\pm 2.47$ & 0.540 $\pm 0.10$  & 0.348 $\pm 0.09$ \\
LLFlow \cite{wang2022low}   &  23.65 $\pm 1.72$ & 0.821 $\pm 0.06$ & 0.159 $\pm 0.04$  & 18.77 $\pm 2.75$  & 0.689 $\pm 0.12$  &  0.305 $\pm 0.08$  & 17.16 $\pm 3.52$  & 0.610 $\pm 0.16$   & 0.441 $\pm 0.12$ \\ 
SDSD  \cite{wang2021seeing}    & 22.63 $\pm 2.10$ & 0.749 $\pm 0.06$ & 0.113 $\pm 0.05$  &  21.22 $\pm 2.06$ & 0.655 $\pm 0.08$ &  0.233 $\pm 0.07$ & 19.93 $\pm 3.38$ & 0.624 $\pm 0.10$ & 0.284 $\pm 0.08$ \\ \hline
\textbf{LF image method} & & & & & & & & & \\ 
LFRetinex \cite{zhang2021effective} & 21.04 $\pm 3.81$  & 0.728 $\pm 0.07$ & 0.186 $\pm 0.07$  & 20.75 $\pm 3.70$  & 0.653 $\pm 0.05$ & 0.256 $\pm 0.05$  & 18.88 $\pm 2.90$  & 0.582 $\pm 0.07$  & 0.357 $\pm 0.08$  \\
L3Fnet \cite{lamba2020harnessing}   & \underline{24.76} $\pm 3.61$  & 0.800 $\pm 0.06$ & 0.064 $\pm 0.02$  & \underline{22.91} $\pm 3.67$  & 0.722 $\pm 0.06$ & \underline{0.105} $\pm 0.03$  & 21.03 $\pm 4.33$  & 0.670 $\pm 0.11$  & 0.169 $\pm 0.06$  \\
FERnet \cite{Lamba2022f}   & 24.62 $\pm 1.51$  & \underline{0.863} $\pm 0.01$  & \underline{0.061} $\pm 0.01$   & 22.77 $\pm 1.23$  & 0.779 $\pm 0.03$ & 0.106 $\pm 0.01$  & 21.48 $\pm 1.10$  & 0.707 $\pm 0.04$  & 0.171 $\pm 0.02$  \\
LRT \cite{zhang2023lrt}   & 24.48 $\pm 2.81$  & 0.750 $\pm 0.06$  & 0.178 $\pm 0.04$  & 22.22 $\pm 2.57$  & 0.678 $\pm 0.06$ & 0.303 $\pm 0.06$  & 20.63 $\pm 3.05$  & 0.611 $\pm 0.07$  & 0.334 $\pm 0.08$  \\
MSPnet \cite{wang2023multi} & 24.42 $\pm 2.70$    & 0.856 $\pm 0.06$  & 0.077 $\pm 0.03$  & 22.72 $\pm 3.33$  & \underline{0.793} $\pm 0.08$ & 0.119 $\pm 0.03$  & \underline{22.63} $\pm 3.34$  & \underline{0.735} $\pm 0.06$ & \underline{0.163} $\pm 0.04$  \\
MSPnet* \cite{wang2023multi} & 25.36 $\pm 2.74$   & 0.866 $\pm 0.06$  & 0.062 $\pm 0.03$  & 23.96 $\pm 3.35$  & 0.797 $\pm 0.08$ & 0.103 $\pm 0.03$  & 22.63 $\pm 3.34$  & 0.735 $\pm 0.06$  & 0.163 $\pm 0.04$  \\
Ours      & \textbf{25.93}$\pm 3.28$   & \textbf{0.871}$\pm 0.05$  & \textbf{0.060}$\pm 0.03$   & \textbf{24.02}$\pm 3.64$   & \textbf{0.810}$\pm 0.05$  & \textbf{0.102}$\pm 0.02$   & \textbf{22.76}$\pm 3.45$   & \textbf{0.741}$\pm 0.07$   & \textbf{0.161}$\pm 0.04$  \\ 
\bottomrule[1.2pt]
\end{tabular}
}
\end{center}
\end{table*}

\section{Experiments}
\label{sec:Experiments}

\subsection{Experiment Settings}
\noindent\textbf{Implementation details}.   
Our proposed method consists of three modules: $\Psi_{i}(\cdot)$ for illumination map estimation, $\Phi_{c}(\cdot)$ for content-associated deep compensation, and $\Omega_{o}(\cdot)$ for deep optimization. We implement these modules by the proposed pseudo-explicit feature interaction with a dense-skip connection. To balance performance and computational expenses, we set the iterative stage number as 3, the layer number as 6, and the channel number as 32 respectively.
During training, we randomly cropped patches of spatial dimensions $32 \times 32$ from LF images. We set the batch size as 2 and initialized the learning rate as $1e^{-4}$, which was halved every 1000 epochs. We used the Adam optimizer \cite{DBLP15} with $\beta_1 = 0.9$ and $\beta_2 = 0.999$. We implemented our method with PyTorch.

\noindent\textbf{Datasets}. Both synthetic LF images from the Inria \cite{shi2019framework} and HCI \cite{honauer2017dataset} benchmarks and real-world LF images captured with a Lytro Illum camera provided by L3F dataset \cite{lamba2020harnessing} were employed to train and test. Specifically, for the synthetic LF dataset, 22 LF images of dimensions $5\times5\times512\times512$ from HCI and 33 LF images of the same dimensions from Inria were selected to form the training set. The test set contains 2 LF images from HCI and 4 LF images from Inria. The data simulation process follows the methodology described in \cite{zhang2023lrt}, employing a low-light factor of $0.2$ and a Gaussian noise deviation of $0.25$. For the L3F dataset \cite{lamba2020harnessing}, we use the same training and test set partitioning provided by the dataset, and the dimensions of the LF data are $7\times7\times434\times625$. 
Note that L3Fnet needs additional SAIs as the input to their view reconstruction block, so the input angular resolution of L3FNet for synthetic LF and real-world LF are $7\times7$ and $9\times9$, respectively.

\begin{figure*}
\begin{center}
\includegraphics[width=1\linewidth]{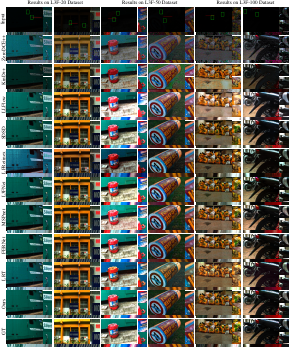}
\end{center}
   \caption{Visual comparisons of enhanced center SAI by different methods on the L3F dataset \cite{lamba2020harnessing}. Selected regions have been zoomed in for better comparison. The EPI and zoomed selected regions of the input image are the original images $\times 10,20,30$ for L3F-20, L3F-50, and L3F-100, respectively. It is recommended to view this figure by zooming in.
   }
\label{fig:l3fresults}
\end{figure*}

\subsection{Comparison with State-of-the-Art Methods}
We compared the proposed DCUNet with three single image enhancement methods, i.e., ZeroDCE++ \cite{li2021learning}, 
KinD++ \cite{zhang2019kindling}, and LLFlow \cite{wang2022low}, 
a recent video enhancement method SDSD \cite{wang2021seeing}, 
and five state-of-the-art LF image enhancement methods, i.e., L3Fnet \cite{lamba2020harnessing}, LFRetinex \cite{zhang2021effective}, MSPnet \cite{wang2023multi}, FERNet \cite{Lamba2022f} and LRT \cite{zhang2023lrt}. For fair comparisons, we retrained all methods with the same dataset and the officially released codes and suggested configurations. The video enhancement method SDSD \cite{wang2021seeing} needs five frames as network input, so we arrange the input sequence of view $(i,j)$ as $[(i-1,j),(i,j-1),(i,j),(i+1,j),(i,j+1)]$ for training and testing. 
We adopted three commonly used metrics PSNR, SSIM, and LPIPS \cite{zhang2018unreasonable} to quantitatively evaluate enhanced image quality.

\noindent\textbf{Quantitative comparisons on the synthetic LF dataset.}
Two strategies were used to simulate the low-light LF images. One is to follow   \cite{lamba2020harnessing} to simulate the data in a fixed illumination environment and use the same photoelectron amplification factor $\alpha = 0.2$, denoted as \textsc{Syn-F}. The other is that the simulation data is in a dynamic illumination environment, and the photoelectron amplification factor changes within a particular range $\alpha \in \left [  0.1,0.3\right ]$, denoted as \textsc{Syn-D}. The standard variance of the adding zero-mean Gaussian noise is 20 and 15, respectively.

Table \ref{tab:my-table1} lists the results of different methods, where it can be observed that:
\begin{compactitem}
    \item our DCUNet outperforms all compared methods on two tasks, achieving 2 dB improvement to the second-best result MSPnet \cite{wang2023multi} on \textsc{Syn-F} and 2.8 dB improvements to the second-best result LFRetinex \cite{zhang2021effective} on \textsc{Syn-D}. 
    Our method has high performance in processing dynamic illumination environment data, which proves the applicability and generalization of the method.
    \item  The performance of ZeroDCE++ \cite{li2021learning} and KinD++ \cite{zhang2019kindling} is inferior to LF-based methods. The reason may be that such single image-based methods cannot utilize multiple SAIs of the LF for noise suppression and illumination enhancement. The SDSD \cite{wang2021seeing} may not efficiently handle LF with significant disparity and severe noise by residual-based center frame alignment. Therefore, the use of partial SAIs information in SDSD \cite{wang2021seeing} offers relatively limited performance improvement. The normalizing flow model employed in LLFlow \cite{wang2022low} demonstrates competitive performance by measuring the visual distance more effectively than using pixel-wise reconstruction loss.
    \item The performance of LF-based methods on \textsc{Syn-D} decreases, compared to \textsc{Syn-F}. But Ours-\textit{D} still achieves much higher quality, demonstrating that our method can effectively enhance dynamic brightness inputs. 
\end{compactitem}

\noindent\textbf{Quantitative comparisons on the L3F dataset.}
We also evaluated the ability of different methods to handle extremely low-light data on the L3F \cite{lamba2020harnessing} dataset, which was captured by the Lytro Illum camera. 
Specifically, L3F-20, L3F-50, and L3F-100 datasets were captured by reducing the exposure time to $20^{-1}$, $50^{-1}$, and $100^{-1}$ of the reference time, respectively. 
As shown in Table \ref{tab:my-table2}, our approach achieves the best performance on all L3F datasets, demonstrating the advantage of the proposed framework to enhance extremely low-light data.

\noindent\textbf{Visual comparisons.}  Fig. \ref{fig:synresults} and Fig. \ref{fig:l3fresults} show the visual comparisons of enhanced LF images by different methods. It can be observed that our method produces better visual results than all the compared methods under all tasks. Specifically, our method can produce
sharper edges at the occlusion boundaries and better high-frequency details at texture regions than other approaches.
Moreover, the visual effects and overall brightness of our results are closest to the ground truth. As shown in Fig. \ref{fig:synresults}, Ours and Ours-\textit{D} can enhance the low-light data with intense noise and obtain clear images. For the LF images with great challenges, such as extremely low light or noise, all the results (including Ours) inevitably suffer from a local color distortion.

\begin{figure*}
\begin{center}
\includegraphics[width=0.90\linewidth]{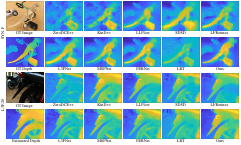}
\vspace{-0.5cm}
\end{center}
   \caption{Visual comparisons of the center depth maps derived by applying the LF depth estimation method in \cite{chen2018accurate} to the enhanced LF images by different methods. Note that the GT depth of \textsc{Syn-F} is available, while the estimated depth of L3F-20 is derived from the GT image.}
\label{fig:depth}
\end{figure*}

\noindent\textbf{Comparisons of the LF parallax structure.}
We also compared the \textit{essential} parallax structures of enhanced LF images by different methods. From Figs. \ref{fig:synresults} and \ref{fig:l3fresults}, it can be seen the formed EPIs from the enhanced LF images by our method show sharper line structures and are closer to ground-truth ones than other methods. 
Moreover, as the precision of the depth map computed from LF images is directly related to the geometric structure underlying the SAIs of an LF image,
we thus compared the depth maps estimated from enhanced LF images of different methods using an identical LF depth estimation method \cite{chen2018accurate}.
As shown in Fig. \ref{fig:depth}, it can be seen that the depth maps of our method are closest to the ground truth, providing sharper edges at occlusion boundaries and preserving smoothness at regions of uniform depth. These above observations demonstrate the advantage of our DCUNet in preserving the parallax structure of enhanced LF images over other methods.

\begin{figure*}
\begin{center}
\includegraphics[width=0.9\linewidth]{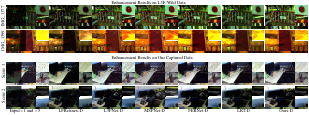}
\vspace{-0.5cm}
\end{center}
   \caption{Comparative visualization of low-light LF image enhancement performed by various methods on the L3F-Wild dataset \cite{lamba2020harnessing} and our dataset, acquired using a Lytro Illum camera. The input SAIs have been amplified by a factor of 5 to enhance detail visibility.} 
\label{fig:illum}
\end{figure*}

\begin{figure}
\begin{center}
\includegraphics[width=0.95\linewidth]{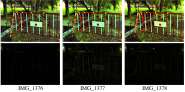}
\vspace{-0.4cm}
\end{center}
   \caption{Results of one scene with different levels of low-light in L3F-Wild dataset \cite{lamba2020harnessing}. The top row is the enhanced result of our DCUNet-\textit{D}, and the bottom row is the central viewpoint of the input LF image.}
\label{fig:dynamic}
\end{figure}

\noindent\textbf{Generalization ability.} To demonstrate the generalization ability, we applied the LF-based methods (i.e., LFRetinex \cite{zhang2021effective}, L3Fnet \cite{lamba2020harnessing}, MSPnet \cite{wang2023multi}, FERNet \cite{Lamba2022f}, LRT \cite{zhang2023lrt} and our DCUNet) trained on \textsc{Syn-D} to two sets of real low-light LF data, one is the L3F-Wild \cite{lamba2020harnessing} dataset and the other is the dataset we acquired using a Lytro Illum camera. 
Fig. \ref{fig:illum} presents the brightness enhancement results across various methodologies.
LFRetinex \cite{zhang2021effective} is observed to cause a noticeable color shift. L3Fnet \cite{lamba2020harnessing} introduces artificial patterns that undermine scene integrity and blur textual details. MSPnet \cite{wang2023multi} enhances overall brightness but suffers from local color shifts and residual noise. FERNet \cite{Lamba2022f} and LRT \cite{zhang2023lrt} exhibit blurring and some loss of structural clarity. In contrast, our proposed method outperforms these approaches by maintaining color fidelity, clarity of structural details, and noise suppression, leading to a more visually appealing enhancement.

In Fig. \ref{fig:dynamic}, we further illustrate the proficiency of our DCUnet in processing images with disparate brightness levels from the L3F-Wild dataset \cite{lamba2020harnessing}. To standardize the input data for a given scene, we employ a straightforward image scaling technique using OpenCV, adjusting the data such that its mean pixel intensity approximates 0.1. The results demonstrate that our method consistently achieves comparable enhancement quality across input scenes with varying levels of brightness.

\begin{figure}
\begin{center}
\includegraphics[width=0.85\linewidth]{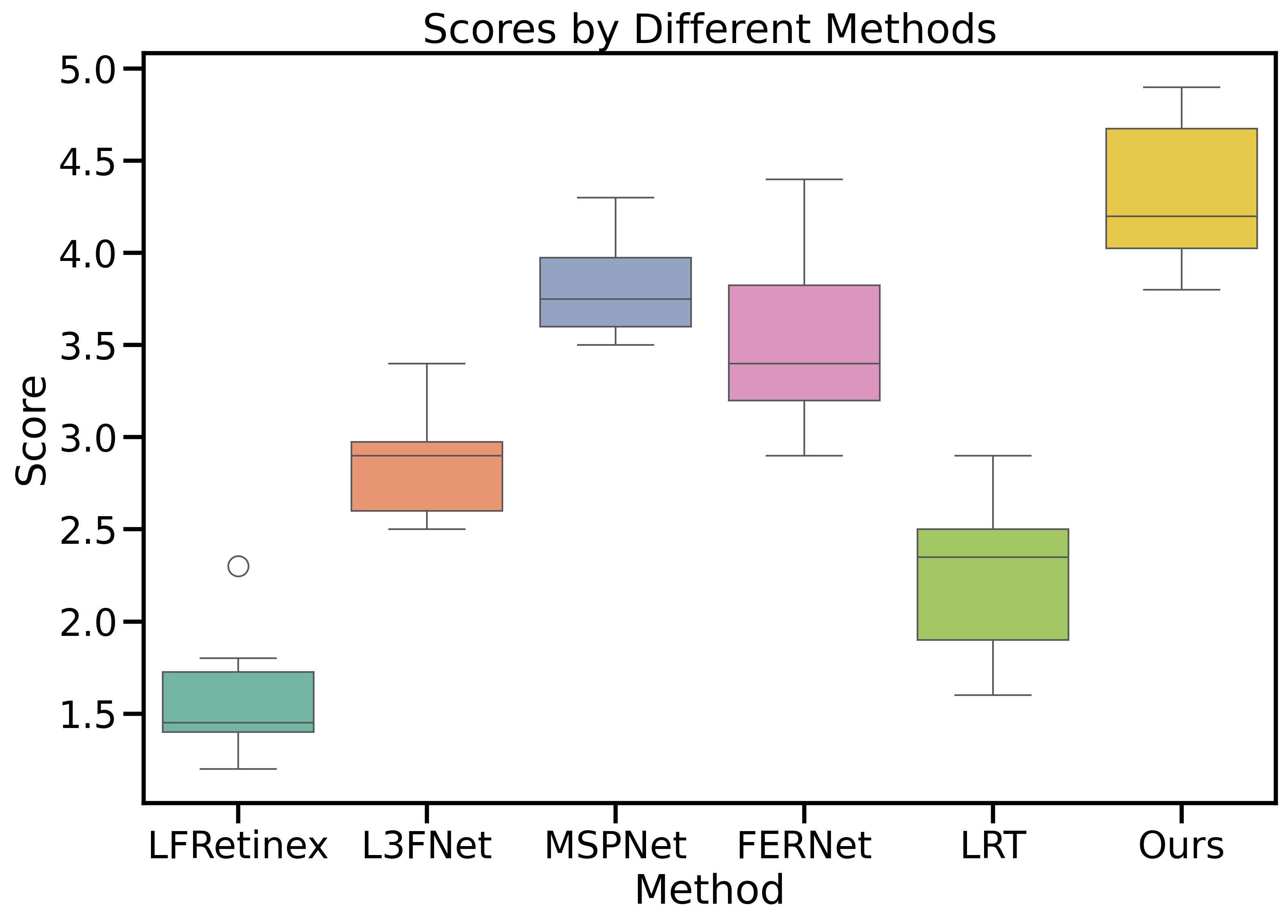}
\vspace{-0.4cm}
\end{center}
   \caption{Boxplot analysis illustrating the distribution of user study scores for LF-based methods across diverse datasets. Scores range from 1 to 5, with higher scores indicative of greater user recognition of image quality.
   }
\label{fig:user-study}
\end{figure}

To quantitatively validate our subjective quality assessment, we conducted a user study with 14 participants, each possessing varied levels of expertise in image quality evaluation. 
For this study, two sets of enhanced images, including central SAI, zoomed-in details, and EPI, were randomly selected from each dataset: \textsc{SYN-F}, L3F \cite{lamba2020harnessing}, L3F-Wild \cite{lamba2020harnessing}, and our own captured LFs.
The participants were requested to assign ratings to the quality of these enhanced images on a 5-point Likert scale, with 1 signifying 'bad', 2: 'poor', 3: 'fair', 4: 'good', and 5: 'excellent'. The evaluative criteria covered several aspects, including brightness, color fidelity, noise, and structural integrity. Note that the participants were blinded to the methods used to generate the results, which were presented in a randomized order.
Statistical analysis of the scores, as shown in Fig. \ref{fig:user-study}, clearly favored our enhancement method, which consistently outperformed other methods.

\begin{table*}[ht]
\caption{Comparisons of inference time (in seconds) and model parameter size (M) of different methods on \textsc{Syn-F} dataset.}
\vspace{-0.3cm}
\centering
\label{tab:time}
\begin{tabular}{c|ccc|c|cccccc}
\toprule[1.2pt] 
 & Zero-DCE++ & KinD++  & LLFlow & SDSD & LFRetinex & L3Fnet & MSPnet & FERnet & LRT & Ours \\ \hline
Time     & 0.06 & 1.26 & 39.55 & 1.38 & 1.01 & 0.66 & 4.99 & 1.41 & 0.21 & 15.79 \\
\#Params & 0.01 & 8.00 & 38.86 & 4.45 & 3.72 & 3.36 & 1.20 & 2.90 & 1.47 & 3.84  \\
\bottomrule[1.2pt]
\end{tabular}
\end{table*}

\noindent\textbf{Computational complexity.} In this study, we conducted a comparative analysis of the inference time and model size of various methods on \textsc{Syn-F} dataset, as presented in Table \ref{tab:time}. All methods were tested on a desktop computer equipped with an Intel CPU i7-8700 @ 3.70GHz, 32 GB RAM, and NVIDIA GeForce RTX 8000. Our findings reveal that the number of parameters in our proposed method is comparable to those of L3Fnet and LFRetinex. However, our method demonstrates a longer inference time, which we attribute to the insufficient utilization of parallel computing capabilities offered by the GPU across the deep network layers. It is worth noting that despite the longer inference time, our approach outperforms other methods in terms of reconstruction quality and noise immunity.

\subsection{Ablation Study}

\begin{table*}
\begin{center}
\caption{Quantitative results of ablation studies assessing the impact of various network architecture components within DCUNet. These components include network modules such as Illumination Map Estimation (IE), Content-Associated Deep Compensation (CDC), and Deep Optimization (DO), as well as DPER structures comprising the Simplified Module (SIM), EPI Boost Structure ($f_{boost}$), and Ray Mixing Structure ($f_{mix}$). ``$\checkmark$"  (resp. ``$\times$") represents the corresponding structure is used (resp. unused). 
}
\vspace{-0.3cm}
\label{tab:ab-structure}
\begin{tabular}{ccc|ccc|cccc}
\toprule[1.2pt]
    \multicolumn{3}{c|}{Network modules} & \multicolumn{3}{c|}{DPEF structures} & \multicolumn{4}{c}{Results} \\ \hline
    IE & CDC & DO & SIM & $f_{boost}$ & $f_{mix}$ & \multicolumn{1}{c}{PSNR$\uparrow$} & \multicolumn{1}{c}{SSIM$\uparrow$} & \multicolumn{1}{c}{LPIPS$\downarrow$} & \#Params(M) \\ \hline
	$\times$ &	$\checkmark$ & $\checkmark$ & $\checkmark$&	$\checkmark$ &  $\checkmark$ & 33.56 & 0.878  & 0.108 & 3.84	\\
    $\checkmark$ &	$\times$ & $\checkmark$ & $\checkmark$&	$\checkmark$ &  $\checkmark$ & 33.21 & 0.871  & 0.116& 3.83\\
	$\checkmark$ &	$\checkmark$ & $\times$ & $\checkmark$&	$\checkmark$ &  $\checkmark$ & 33.50 & 0.879  & 0.110 & 3.85 \\ \hline
	$\checkmark$ &	$\checkmark$ & $\checkmark$ & $\checkmark$&	$\times$ & $\times$     & 33.37 & 0.875  & 0.112  & 3.95	\\
    $\checkmark$ &	$\checkmark$ & $\checkmark$ & $\checkmark$&	$\times$ & \checkmark   & 33.64 & 0.884  & 0.104  & 3.86	\\
	$\checkmark$ &	$\checkmark$ & $\checkmark$ & $\checkmark$&	$\checkmark$ & $\times$ & 33.72 & 0.884  & 0.104  & 3.90\\
	$\checkmark$&	$\checkmark$ & $\checkmark$ & $\checkmark$&	$\checkmark$ & $\checkmark$ & \textbf{33.96} & \textbf{0.886}  & \textbf{0.103} & 3.84 \\
\bottomrule[1.2pt]
\end{tabular}
\end{center}
\end{table*}

\noindent\textbf{Network modules and DPEF structures.}
The detailed ablation study results for the proposed DCUNet are presented in Table \ref{tab:ab-structure}, which includes the examination of the main network modules and the various configurations of the proposed DPEF for low-light LF enhancement. To ascertain the contribution of each network module, i.e., Illumination Map Estimation (IE), Content-Associated Deep Compensation (CDC), and Deep Optimization (DO), we substituted the specifically tailored modules with a sequence of DPEF units, calibrated to preserve a comparable number of parameters. The results, as indicated in Table \ref{tab:ab-structure}, demonstrate that the full configuration of DCUNet yields the highest performance metrics, thereby confirming the efficacy of each dedicated network module.

The pseudo-explicit feature interaction is designed for effective feature extraction of low-light LFs without depth information by converging rays and boosting EPI convolution. To validate the efficacy of the proposed structures $f_{boost}$ and $f_{mix}$, we incrementally introduced them into the simplified module (SIM) base configuration and adjusted feature channel numbers to align the overall parameter. The ablation study results, as delineated in Table \ref{tab:ab-structure}, reveal that both $f_{boost}$ and $f_{mix}$ contribute uniquely to the feature extraction process. It is evident that the full deployment of the proposed DPEF achieves the best performance. 

To demonstrate the superiority of our proposed pseudo-explicit feature interaction, we performed comparative experiments by replacing our module with Spatial-Angular Separable convolution \cite{yeung2018light}, Spatial-Angular Versatile (SAV) convolution \cite{cheng2022spatial} and Disentangling convolution \cite{wang2022disentangling}. For clarity, we refer to these variants as Ours-SAS, Ours-SAV and Ours-Distg, respectively. We adjusted the number of feature channels to align the overall parameter count across models for a fair comparison. The strategy of full-angular convolution combined with $1\times1$ convolution to generate all angular information in Disentangling convolution may not be suitable for processing low-light light field images with large baseline and severe noise. The results presented in Table \ref{tab:abfeature} indicate that our DCUNet, incorporating the proposed pseudo-explicit feature interaction with specialized techniques such as EPI boost convolution and ray convergence, tailored for low-light LF enhancement, achieves superior performance over the Ours-SAS, Ours-SAV, and Ours-Distg models. This clearly validates the benefits of our pseudo-explicit feature interaction approach. 

\begin{table}
\caption{Quantitative results of ablation study about different feature extraction methods on \textsc{Syn-F}.}
\vspace{-0.3cm}
\begin{center}
\begin{tabular}{c|cccc}
\toprule[1.2pt]
Method     & PSNR $\uparrow$ & SSIM $\uparrow$ & LPIPS $\downarrow$ & \#Params(M) \\ \hline 
Ours-SAS       & 32.38 & 0.853  & 0.121  &  4.02  \\
Ours-SAV       & 32.09 & 0.852  & 0.135 & 4.05 \\
Ours-Distg     & 32.74 & 0.860  & 0.129 & 3.62 \\
Ours           & \textbf{33.96} & \textbf{0.886}  & \textbf{0.103} & 3.84  \\ 
\bottomrule[1.2pt]
\end{tabular}
\end{center}
\label{tab:abfeature}
\end{table}

\begin{table}
\caption{Comparisons of different numbers of pseudo-explicit feature interaction layers in dense-skip connection and different numbers of iterative stages.}
\vspace{-0.3cm}
\begin{center}
\begin{tabular}{c|c|cccc}
\toprule[1.2pt]
Fixed        & Dynamic    & PSNR $\uparrow$ & SSIM $\uparrow$ & LPIPS $\downarrow$ & \#Params(M) \\ \hline 
\multirow{4}{*}{$S=3$} 
& $L=2$    & 33.02 & 0.866  & 0.117 & 1.26  \\
& $L=4$    & 33.59 & 0.878  & 0.105 & 2.54     \\
& $L=6$    & 33.96 & 0.886  & 0.103 & 3.84  \\
& $L=8$    & \textbf{34.02} & \textbf{0.890}  & \textbf{0.102} & 5.20  \\ \hline
\multirow{4}{*}{$L=6$} 
& $S=1$    & 33.34 & 0.873  & 0.111 & 1.28  \\
& $S=2$    & 33.67 & 0.881  & 0.105 & 2.57     \\
& $S=3$    & 33.96 & 0.886  & 0.103 & 3.84  \\
& $S=4$    & \textbf{34.15} & \textbf{0.890}  & \textbf{0.101}  & 5.14 \\ 
\bottomrule[1.2pt]
\end{tabular}
\end{center}
\label{tab:my-table3}
\end{table}

\begin{figure}
\begin{center}
\includegraphics[width=0.95\linewidth]{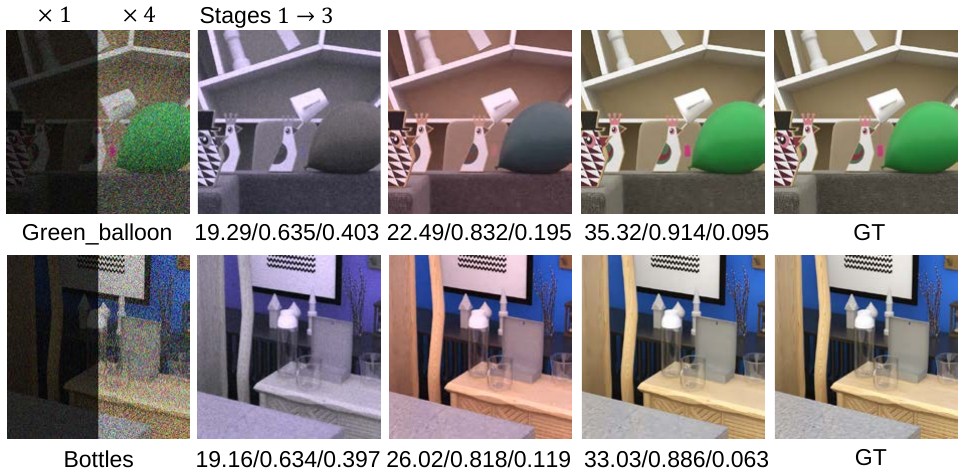}
\vspace{-0.4cm}
\end{center}
   \caption{Visualization of the intermediate output for a configuration of $S=3$ and $L=6$ at each stage of our DCUNet. We multiply the central SAI by $\times4$ to show the severe noise. The PSNR/SSIM/LPIPS values are listed below.}
\label{fig:stages}
\end{figure}

\noindent\textbf{The numbers of iterative stages and convolutional
layers.} To analyze the impact of stage $S$ and layer $L$ numbers on performance, we compared DCUNet results with different configurations, i.e., $S=$ 1, 2, 3, and 4, $L=$ 2, 4, 6, and 8. Taking the \textsc{Syn-F} dataset as an example, we separately carried out ablation studies on
these two factors. 
By analyzing the results in Table \ref{tab:my-table3}, we can observe that the PSNR/SSIM values gradually improve from 1 to 4 stages and 2 to 8 layers, and the LPIPS values demonstrate the inverse tendency. In practice, we chose 3 stages and 6 layers in our framework to trade off LF enhancement quality and computational costs. 

Furthermore, we visualize the output of each stage of DCUNet in Fig. \ref{fig:stages} to illustrate that the deep unfolding optimization procedure is effective. We can observe that the enhancement quality gradually improves from the first stage to the last one, and the noise is suppressed in the first stage.

\begin{figure}
\begin{center}
\includegraphics[width=0.9\linewidth]{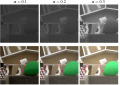}
\vspace{-0.4cm}
\end{center}
   \caption{Estimated illumination maps and enhancement results for input low-light LF images with amplification factor $\alpha =$ 0.1, 0.2, and 0.3.
   }
\label{fig:imap}
\end{figure}

\noindent\textbf{Independently optimized illumination map.}
To demonstrate the effectiveness of our signal-dependent illumination map estimation strategy, we conducted an ablation study by independently optimizing the enhancement results and illumination maps, akin to the methodology by Wu et al. \cite{wu2022uretinex}. Specifically, we replaced the coarse illumination map estimation and subsequent closed-form solution with an optimization approach similar to that presented in Section \ref{sec:Method}. In this approach, we fixed the enhancement result $\textbf{L}_{n}^{k}$, and the illumination map was optimally solved. We denote this modified model as Ours-Dual. We evaluated the performance of both strategies on all datasets and presented the results in Table \ref{tab:ab-dual}, where the superior performance of Ours to Ours-Dual indicates the efficacy of the used signal-dependent illumination map estimation strategy. Moreover, this strategy facilitates a more concise model architecture. 

Additionally, we have provided visual representations of the estimated illumination maps for input low-light LF images, subject to amplification factors $\alpha =$ 0.1, 0.2, and 0.3, as depicted in Fig. \ref{fig:imap}. We can observe that the estimated illumination maps positively correlate with the brightness of the input low-light LF images. This correlation substantiates the proficiency of our IE module in adapting to and processing input LF images across a spectrum of brightness levels.

\begin{table*}
\begin{center}
\caption{Quantitative comparisons (PSNR/SSIM/LPIPS) of different structures for illumination map estimation.}
\vspace{-0.3cm}
\label{tab:ab-dual}
\begin{tabular}{c|ccc|ccc|ccc|ccc}
\toprule[1.2pt]
\multirow{2}{*}{} & \multicolumn{3}{c|}{Syn-F} & \multicolumn{3}{c|}{L3F-20} & \multicolumn{3}{c|}{L3F-50} & \multicolumn{3}{c}{L3F-100} \\ \cline{2-13}
 & PSNR $\uparrow$  & SSIM $\uparrow$   & LPIPS $\downarrow$   & PSNR $\uparrow$    & SSIM $\uparrow$   & LPIPS $\downarrow$   & PSNR $\uparrow$  & SSIM $\uparrow$   & LPIPS $\downarrow$  & PSNR $\uparrow$  & SSIM $\uparrow$   & LPIPS $\downarrow$ \\ \hline
Ours-Dual & 32.41 & 0.852 & 0.127 & 25.44 & 0.862 & 0.066  & 23.86 & 0.805 & 0.111  & 22.73 & 0.719 & 0.175   \\
Ours      & \textbf{33.96} & \textbf{0.886} & \textbf{0.103} & \textbf{25.93}   & \textbf{0.871}  & \textbf{0.060}   & \textbf{24.02}   & \textbf{0.810}  & \textbf{0.102}   & \textbf{22.76}   & \textbf{0.741}   & \textbf{0.161}  \\ 
\bottomrule[1.2pt]
\end{tabular}
\end{center}
\end{table*}

\begin{table}
\caption{Quantitative results of the ablation studies for different loss settings. ``$\checkmark$"  (resp. ``$\times$") represents the corresponding loss component is used (resp. unused).}
\vspace{-0.4cm}
\begin{center}
\begin{tabular}{ccc|ccc}
\toprule[1.2pt]
    \texttt{$L_1$} & $\texttt{SSIM}(\cdot)$ & $\texttt{Per}(\cdot)$ & \multicolumn{1}{c}{PSNR$\uparrow$} & \multicolumn{1}{c}{SSIM$\uparrow$} & \multicolumn{1}{c}{LPIPS$\downarrow$}  \\ \hline
	$\checkmark$&	$\times$ & $\times$   & 33.67 & 0.879  & 0.106	\\
	$\checkmark$&	$\checkmark$&$\times$ & 33.83 & 0.885  & 0.104	\\
	$\checkmark$&	$\checkmark$&$\checkmark$ & \textbf{33.96} & \textbf{0.886}  & \textbf{0.103} \\
\bottomrule[1.2pt]
\end{tabular}
\end{center}
\label{tab:table-loss}
\end{table}

\noindent\textbf{Loss configurations.} We investigated the effectiveness of different loss components, namely L1 loss $L_1$, SSIM loss $\texttt{SSIM}(\cdot)$, and perceptual loss $\texttt{Per}(\cdot)$. The results in Table \ref{tab:table-loss} demonstrate a gradual performance improvement, indicating that incorporating multiple loss components could enhance our DCUNet's performance.

\begin{table}
\caption{Comparisons of inference time (in seconds) and model parameter size (M) of different strategies on \textsc{Syn-F} dataset.} 
\vspace{-0.4cm}
\begin{center}
\begin{tabular}{c|ccc|cc}
\toprule[1.2pt]
Method     & PSNR $\uparrow$ & SSIM $\uparrow$ & LPIPS $\downarrow$  & \#Params & Time \\ \hline 
MSPnet \cite{wang2023multi}  & 31.93 & 0.860 & 0.131 & 4.99 & \textbf{1.20} \\
Ours-Share     & 30.16 & 0.804  & 0.179 & 1.29 & 15.55 \\
Ours-C64       & 33.21 & 0.877  & 0.112 & 3.14 & 3.91 \\
Ours           & \textbf{33.96} & \textbf{0.886}  & \textbf{0.103} & 3.84 & 15.79  \\ 
\bottomrule[1.2pt]
\end{tabular}
\end{center}
\label{tab:ab-limit}
\end{table}

\subsection{Limitation and Discussion}
We recognize that our proposed method encounters limitations in harnessing the full potential of parallel computing architectures, primarily due to the adoption of a deeper network structure. To address this shortcoming, we attempted to flatten the network structure to facilitate parallel computing. This involved increasing the number of feature channels from 32 to 64 while concurrently diminishing the number of processing stages to $S=1$ and the number of layers to $L=4$. We refer to this modified model as Ours-C64 and present its performance, inference time, and parameter size in Table \ref{tab:ab-limit}. Our findings suggest that, under similar parameter quantities, the speed of the whole framework is significantly accelerated, while the PSNR is still 1.3dB higher than that of MSPnet \cite{wang2023multi}. 

Furthermore, parameter sharing across different stages of optimization is conventionally acknowledged as a viable strategy for minimizing the model's parameter burden. Pursuant to this understanding, we applied parameter sharing within the optimization phase, an approach we refer to as Ours-Share. However, the comparative analysis revealed that Ours-Share suffers a marked performance degradation relative to the original Ours model. This suggests that applying parameter sharing in the optimization phase may constrain the solution space excessively, rendering it unsuitable for our proposed method.

\section{Conclusion}
\label{sec:Conclusion}

We have presented a novel learning-based method for enhancing low-light LF images. Specifically, we concentrated on utilizing LF properties to tackle extremely low-light enhancement with severe noise. As the basic module of our method, the pseudo-explicit feature interaction can effectively extract the features of low-light LF images. 
Our DCUNet can enhance 4D LF from extremely low-light data with severe noise using the proposed illumination estimation, content-associated deep compensation, and deep optimization modules and achieve substantially higher quality than state-of-the-art approaches.

\small
\bibliographystyle{IEEEtran}
\bibliography{egbib}

\begin{IEEEbiography}[{\includegraphics[width=1in, height=1.25in, clip, keepaspectratio]{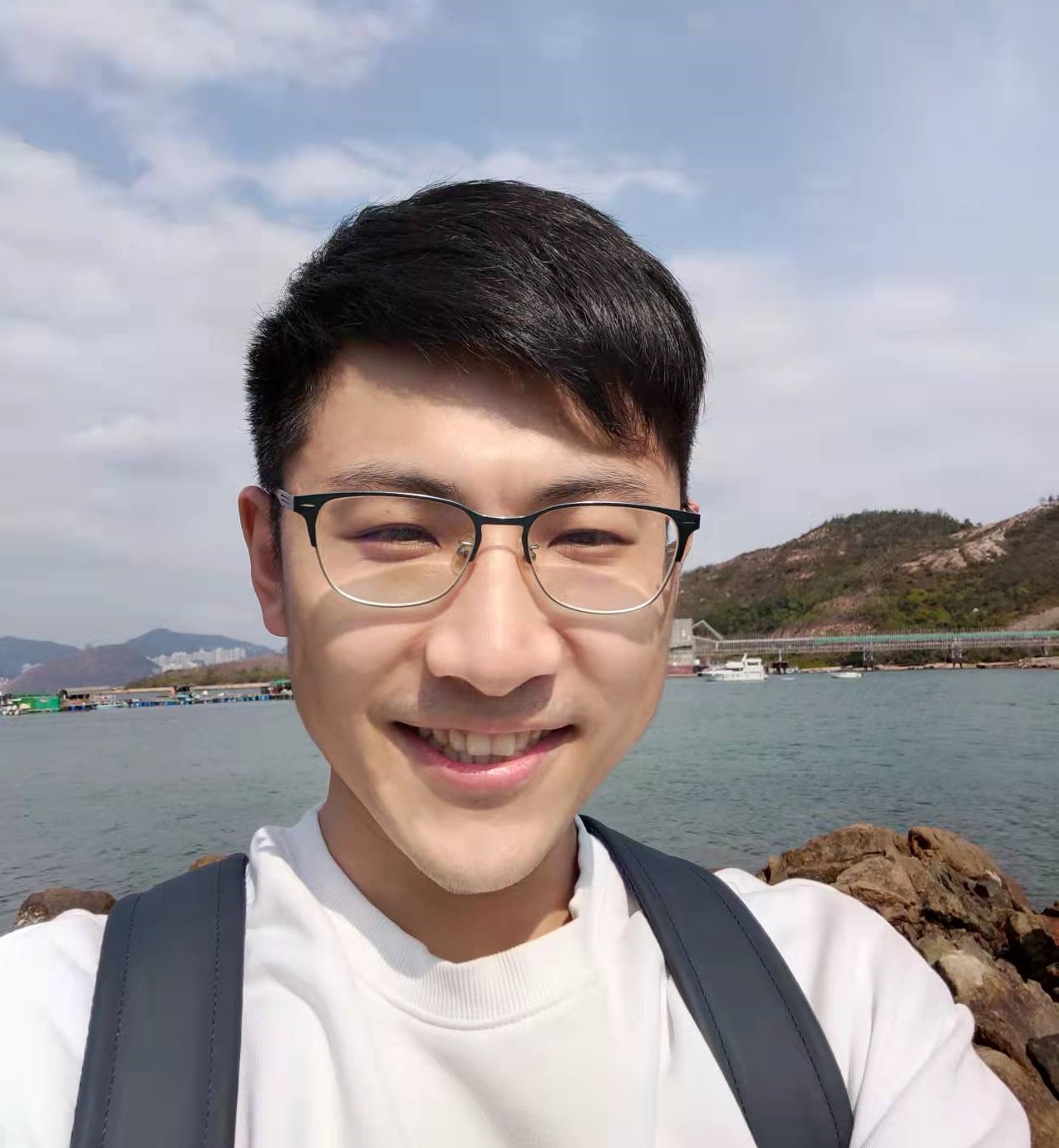}}]{Xianqiang Lyu} received B.Eng. and M.Eng. degrees from the School of Computer Science, Northwestern Polytechnical University, Xi’an, China, in 2017 and 2020, respectively. He is currently pursuing a Ph.D. degree with the Department of Computer Science, City University of Hong Kong. His research interests include computational photography, light field image processing, neural rendering, and deep learning.
\end{IEEEbiography}

\begin{IEEEbiography}[{\includegraphics[width=1in, height=1.25in, clip, keepaspectratio]{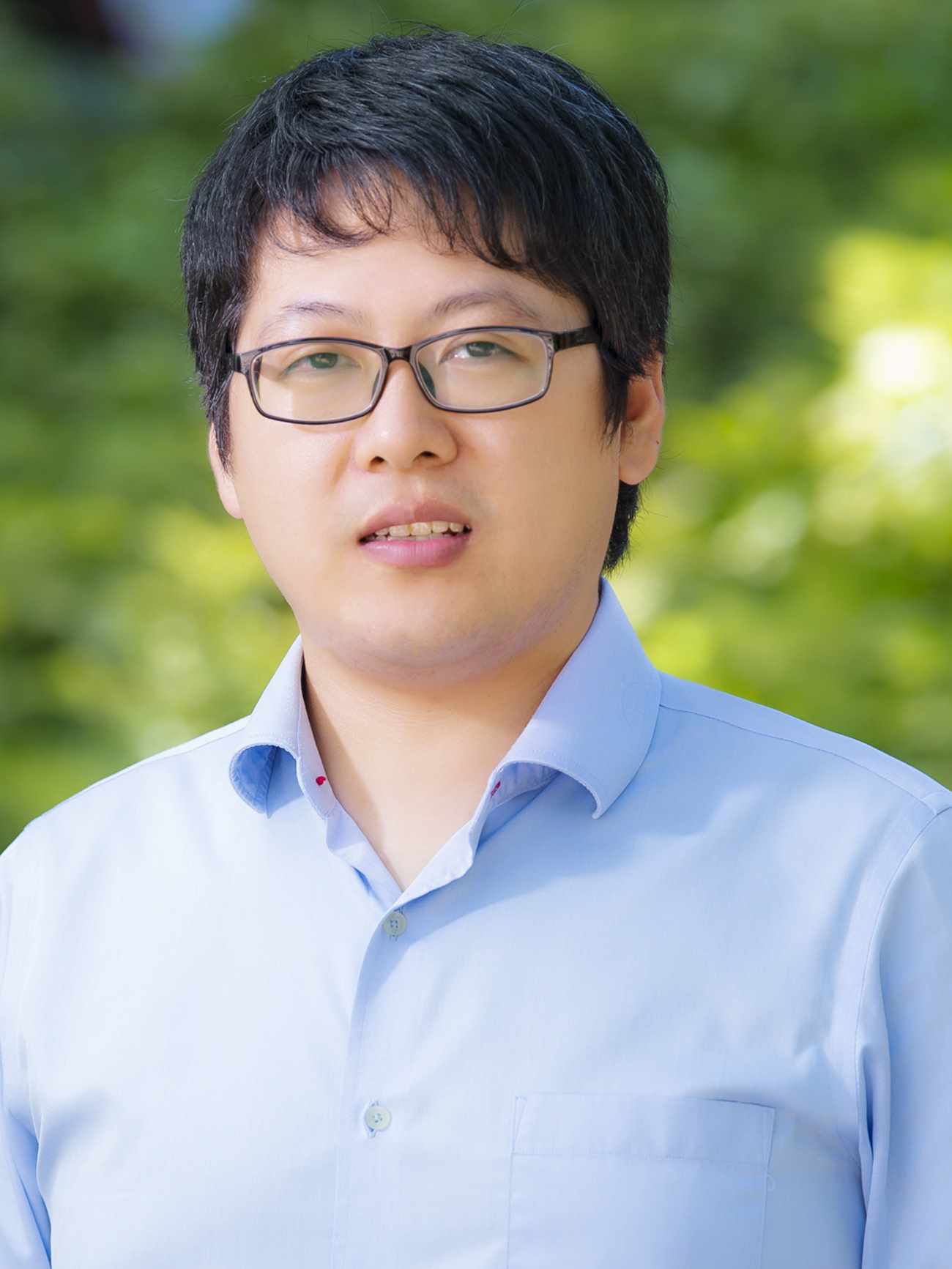}}]{Junhui Hou} (Senior Member, IEEE)  is an Associate Professor with the Department of Computer Science, City University of Hong Kong. He holds a B.Eng. degree in information engineering (Talented Students Program) from the South China University of Technology, Guangzhou, China (2009), an M.Eng. degree in signal and information processing from Northwestern Polytechnical University, Xi’an, China (2012), and a Ph.D. degree from the School of Electrical and Electronic Engineering, Nanyang Technological University, Singapore (2016). His research interests are multi-dimensional visual computing.

He received the Early Career Award (3/381) from the Hong Kong Research Grants Council in 2018. He is an elected member of IEEE MSATC, VSPC-TC, and MMSP-TC. He has served or is serving as an Associate Editor for \textit{IEEE Transactions on Visualization and Computer Graphics}, \textit{IEEE Transactions on Image Processing}, \textit{IEEE Transactions on Circuits and Systems for Video Technology}, \textit{Signal Processing: Image Communication}, and \textit{The Visual Computer}.

\end{IEEEbiography}

\end{document}